# Flight Trajectory Prediction Using an Enhanced CNN-LSTM Network

Qinzhi Hao[1], Jiali Zhang[2], Tengyu Jing[2], Wei Wang[2,*]
[1] School of Aeronautical Engineering, Air Force Engineering University, Xi'an 710051, China
[2] School of Information Engineering, Xidian University, Xi'an 710071, China
[*] Corresponding author: wwang@mail.xidian.edu.cn

*Abstract- Aiming at the problem of low accuracy of flight trajectory prediction caused by the high speed of fighters, the diversity of tactical maneuvers, and the transient nature of situational change in close range air combat, this paper proposes an enhanced CNN-LSTM network as a fighter flight trajectory prediction method. Firstly, we extract spatial features from fighter trajectory data using CNN, aggregate spatial features of multiple fighters using the social-pooling module to capture geographic information and positional relationships in the trajectories, and use the attention mechanism to capture mutated trajectory features in air combat; subsequently, we extract temporal features by using the memory nature of LSTM to capture long-term temporal dependence in the trajectories; and finally, we merge the temporal and spatial features to predict the flight trajectories of enemy fighters. Extensive simulation experiments verify that the proposed method improves the trajectory prediction accuracy compared to the original CNN-LSTM method, with the improvements of 32% and 34% in ADE and FDE indicators.*

## I. INTRODUCTION

The basic principle of flight trajectory prediction is to utilize the spatial position information of the historical trajectory of the fighter and predict the position of the fighter at a certain moment in the future according to the current flight status and the tactical system of the many-to-many air war. Accurate prediction of the flight trajectory of the enemy aircraft can prejudge the movement of the enemy aircraft, and preempt the formation of the observe, orient, decide, act (OODA) cycle [1]. In addition, better prediction of the position information of the aircraft allows our fighters to make a more accurate and timely response, to effectively defend and intercept the enemy fighters.

Recurrent Neural Networks (RNN) and Long Short Term Memory (LSTM) can learn relatively well for data with temporal features and are widely used in trajectory prediction. For example, Alahi et al [2] proposed Social for LSTM (Social-LSTM), which simulates interactions between pedestrians by constructing a social pooling layer. Since Convolutional Neural Network (CNN) have the excellent property of extracting image features, Mottaghi et al [3] used CNNs and RNNs together to accomplish the task of predicting the motion trajectory of an object after a force is applied to it. Over time, CNNs have significantly outperformed RNNs in the time-series domain with astounding results [4].

With the development of time, it has been found that the attention mechanism can significantly improve the accuracy of trajectory prediction models [5-8]. In 2019, Haddad et al [9] proposed a pedestrian trajectory prediction model that considers contextual information. The model consists of three parts: a spatial attention model, a temporal attention model, and a motion model. Fernando et al [10] proposed a human trajectory prediction method with LSTM and soft and hard connective attention mechanisms. This attention mechanism can dynamically adjust the prediction results of the model, thus improving the accuracy and robustness of the prediction.

The successful application of deep learning in pedestrian trajectory prediction provides new ideas for fighter flight trajectory prediction in the military field. Zhang Hongpeng et al [11] used gated recursive units to predict the air combat trajectories of fighters with frequently changing maneuvers. To further improve the accuracy of air combat trajectory prediction, they again used a network with CNN architecture to predict air combat trajectories in 3D space [12]. Zhang, A. et al [13] proposed an attention-based convolutional long and short-term memory network, which treats the trajectory prediction problem as a categorization problem and calculates the probability that the target aircraft is reachable to each space in the region.

In the face of the problem of poor flight trajectory prediction for fighters, this paper combines CNN and LSTM and introduces an attention mechanism to deal with the spatiotemporal features of flight trajectories in a more detailed and dynamic method. Through tests on the corresponding datasets, this paper verifies the effectiveness and feasibility of the proposed method in fighter flight trajectory prediction, achieving remarkable prediction results with high efficiency and low error, and providing a new paradigm for improving the accuracy and robustness of flight trajectory prediction.

## II. CNN-LSTM NETWORK WITH ATTENTION MECHANISM AND SOCIAL-POOLING

### A. Model Introduction

The traditional trajectory prediction model with statistics combined with expert strategy has the problems of incomplete trajectory information feature extraction and over-reliance on the pilot's flight experience, which makes the method's accuracy relatively poor. The trajectory prediction based on machine learning algorithm improves the prediction accuracy but also has the following difficulties:

1) How to extract the temporal and spatial characteristics of the historical trajectory of a fighter;

2) How to model the interplay of competition and collaboration that exists between fighters;

3) How to deal with the presence of sudden trajectory changes during fighter confrontation.

This paper uses a CNN-LSTM network structure for spatio-temporal feature extraction. Due to the high degree of freedom in 3D space, spatial feature extraction is difficult, so we borrow the idea of social attributes from the Social-LSTM network, use the social-pooling module to aggregate the spatial features of multiple fighters, and model the relationship between confrontation and collaboration between fighters. This paper introduces an attention mechanism to focus on the points where trajectories undergo mutations. When outputting the predicted trajectories, we adopt the sliding window method: generating a subsequent new prediction point for every neighboring 8 points and gradually calculating the predicted position information of the future 8 points from the input 8 points.

### B. Overall Modeling Framework

As shown in Fig. 1, the overall network in this paper adopts an encoder-decoder structure. On the encoder side, the CNN network is used to extract features from the $traj = \{traj_1, traj_2, traj_3, \cdots, traj_n\}$ historical trajectories of the fighters, obtain the respective hidden layer features of the current fighters $h_i$, and send the features to the attention mechanism module and social-pooling module respectively in parallel. The spatial features in the historical trajectories of the fighters themselves are extracted through the attention mechanism module, and the spatial features among different fighters are aggregated through the social-pooling module, and then features are aggregated and sent to the LSTM network for temporal feature extraction, and finally all the features are sent to the decoder, which outputs all the predicted trajectories of the fighters $traj_p$.

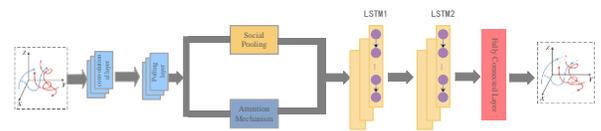

Fig. 1 CNN-LSTM Network Prediction Model With Attention Mechanism and Social-Pooling

### a). CNN Module

The historical trajectory information of the fighter $traj = \{traj_1, traj_2, traj_3, \cdots, traj_n\}$ is taken as input and fed into the CNN layer. This model adopts one-dimensional convolution, which will be the convolution kernel only for the temporal dimension of the input sequence to perform convolution operation, without considering the spatial dimension. The weight matrix $W_i$ is a $3 \times m$ learnable matrix of real numbers and $m$ is the number of features. Feature extraction is performed once on the sequence vector at each time step to obtain the $i$ th fighter jet feature $o_i$, which is calculated as follows:

$$o_i = f(W_i * traj_i + b_i) \quad (1)$$

where $f$ is the nonlinear activation function, $b_i$ is the bias, and $traj_i$ is the trajectory information of the $i$ th fighter plane at time $t : t+7$. The $n$ flight trajectory features obtained by the convolution operation are fed into a pooling layer of size and step size 2 for maximum pooling operation to obtain $n$ feature maps.

*b). Attention Mechanism Module*

The attention mechanism chosen in this paper is the improved Luong Attention, the original structure of Luong Attention is shown in Fig. 2:

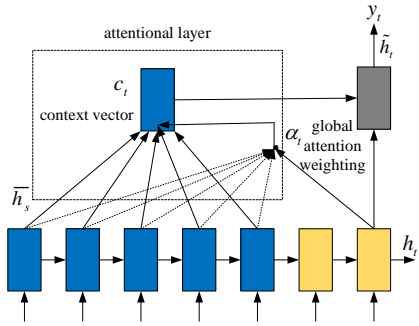

Fig. 2 Luong Attention Implementation Diagram

The computation of $\alpha_t$, $c_t$, $\tilde{h}_t$ in Fig.2 is the same as in [14].

The final output $y_t$ is calculated as follows:

$$y_t = W_h \tilde{h}_t + b_h \quad (2)$$

Unlike the traditional attention mechanism module which is only used at the decoder side, this paper introduces the attention mechanism module at the encoder side and hence modifies Luong Attention to extract features and extract the dependency between long-term sequences. The attention mechanism adopted in this paper assigns a higher weight to a segment of the sequence where the trajectory mutates, enabling the network to better predict complex and variable trajectories.

First, the feature $o_i^t$ of the $i$ th fighter at moment t output from the CNN module is computed with the hidden state $h_i^{t-1}$ of the attention mechanism module at the previous moment to obtain the hidden state of the attention module at the current moment, which is given by Eq:

$$h_i^t = f(h_i^{t-1}, o_i^t) \quad (3)$$

where $f$ is the feature extraction function, in this paper, we choose the LSTM network for adaptive feature extraction.

In addition, each feature of $o_i^t$ is assigned a certain attention weight $\alpha_k^t$, denoted as the importance of the kth feature at moment t, and $l_t^1, l_t^2, \cdots, l_t^m$ is the feature element of $o_i^t$. After the assignment $o_i$ is updated as:

$$\tilde{o}_i^t = (\alpha_1^t l_t^1, \alpha_2^t l_t^2, \cdots, \alpha_t^m l_t^m) \quad (4)$$

$$\alpha_k^t = \frac{exp(e_k^t)}{\sum_{k=1}^{m}(e_k^t)} \quad (5)$$

$$e_k^t = v_e^T tanh(W_e[h_{t-1}, s_{t-1}] + U_e o_k^t) \quad (6)$$

where $[h_{t-1}, s_{t-1}]$ is the hidden state and cell state in the ff feature extraction function, respectively; $v_e$, $W_e$, and $U_e$ are all learnable parameter matrices.

Finally, Eq. (3) is updated as:

$$h_i^t = f(h_i^{t-1}, \tilde{o}_i^t) \quad (7)$$

*c). Social-Pooling Module*

To characterize the "social" relationship between fighters, this paper introduces social-pooling for information fusion. The structure is shown in Fig. 3, where the black points are the input fighter and the other points are the fighters around the fighter.

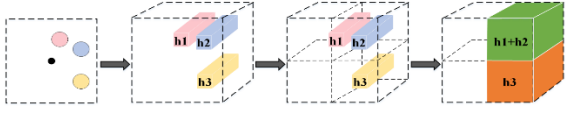

Fig. 3 Social-Pooling Structure Diagram

The features extracted through the CNN layer cannot capture the spatial interactions between the fighters. Therefore we address this limitation by connecting the CNN layer to the LSTM layer through a social-pooling operation. Unlike traditional CNN-LSTM networks, social-pooling allows spatially neighboring features to share information. The formula is as follows:

$$H = \sum_{j \in N_i} 1_{lmn}\left[x_t^j - x_t^i, y_t^j - y_t^i, z_t^j - z_t^i\right] h_{t-1}^j \quad (8)$$

where $h_{t-1}^j$ denotes the hidden state of the previous layer of the node, and $1_{lmn}$ is an indicator function indicating whether the node $[x, y, z]$ is in the lattice $[l, m, n]$ or not, and 1 if it is in the lattice, and 0 otherwise. is the set of "neighbors" of the first node. If it is in the grid, it is 1, otherwise, it is 0. $N_i$ is the set of "neighbors" of the $i$ th node.

*d). LSTM Module*

Through the above methods, the network has been able to better extract spatial features in trajectory information. To better extract time features, this paper introduces the LSTM module, which utilizes the sensitivity of the LSTM network to temporal information, learns the temporal features of the historical trajectory information, and then updates the initial parameters of the network, so that the overall network can better extract the temporal and spatial dimensional features, and can better predict the future trajectories.

The spatial features extracted above are spliced and merged to obtain the state features of the hidden layer:

$$x = \{x_1, x_2, \cdots, x_n\} \quad (9)$$

where $n$ is the number of fighters.

The state information $h_i$ of the fighter at the current moment is obtained by LSTM:

$$h_i = LSTM\left(x_i^t, x_i^{t+1}, x_i^{t+2}, x_i^{t+3}, \cdots, x_i^{t+7}\right) \quad (10)$$

where $x_i^t$ is the hidden state of the $i$ th fighter at the moment $t$.

*e). Trajectory Prediction Output Module*

The predicted trajectory output module aggregates the previously obtained features and finally outputs the predicted trajectory $traj_p$ of the fighter from the fully connected layer.

## III. EXPERIMENT AND RESULT ANALYSIS

*A. Model Introduction*

In this paper, we use the Python language and PyTorch framework for model construction and simulation experiments and adopt a dual-card crossfire 1080ti graphics card host as the experimental hardware environment.

*B. Introduction to the Dataset*

In this paper, air combat simulation is performed on DCS World platform to record the flight data such as altitude, speed, longitude, latitude, roll angle, yaw angle, and pitch angle of the fighter. Subsequently, the recorded data were imported into the acmi file, and finally, the acmi file was run in the tacview software to export the flight data in the form of a csv file. After the obtained data were low-pass filtered to filter out the noise interference they are divided into a training set and a test set in the ratio of 4:1.

*C. Evaluation Indicators*

1) ADE(Average Displacement Error): Used to measure the average error of the trajectory prediction model throughout the prediction process, calculated as follows:

$$ADE = \frac{1}{n}\sum_{i=1}^{n}\frac{1}{t_{pred}}\sum_{t=t_{obs}+1}^{t_{obs}+t_{pred}}\sqrt{\begin{array}{l}\left(x_i^t - \tilde{x}_i^t\right)^2 + \\ \left(y_i^t - \tilde{y}_i^t\right)^2 + \left(z_i^t - \tilde{z}_i^t\right)^2\end{array}} \quad (11)$$

where the dataset contains a total of $n$ moments; $t_{obs}$ represents the trajectory data containing $obs$ historical moments during the prediction process; $x_i^t$, $y_i^t$, and $z_i^t$ represent the Cartesian coordinate values (true values) of fighter $i$ at moment $t$, while $\tilde{x}_i^t$, $\tilde{y}_i^t$, and $\tilde{z}_i^t$ represent the Cartesian coordinate values (predicted values) of fighter $i$ predicted by the network at moment $t$, respectively.

2) FDE(Final Displacement Error): the Euclidean distance between the endpoint of the predicted trajectory (i.e., the last moment $t_{pred}+t_{obs}$) and the endpoint of the true trajectory. Its calculation formula is as follows:

$$FDE = \sqrt{\begin{aligned}&\left(x_i^{t_{obs}+t_{pred}} - \tilde{x}_i^{t_{obs}+t_{pred}}\right)^2 + \\ &\left(y_i^{t_{obs}+t_{pred}} - \tilde{y}_i^{t_{obs}+t_{pred}}\right)^2 + \\ &\left(z_i^{t_{obs}+t_{pred}} - \tilde{z}_i^{t_{obs}+t_{pred}}\right)^2\end{aligned}} \quad (12)$$

where $x_i^{t_{obs}+t_{pred}}$, $y_i^{t_{obs}+t_{pred}}$ and $z_i^{t_{obs}+t_{pred}}$ represent the values of the Cartesian coordinates of fighter $i$ at the last moment (true values), while $\tilde{x}_i^{t_{obs}+t_{pred}}$, $\tilde{y}_i^{t_{obs}+t_{pred}}$ and $\tilde{z}_i^{t_{obs}+t_{pred}}$ represent the values of the Cartesian coordinates of fighter $i$ at the last moment predicted by the network (predicted values), respectively.

In addition to the two errors mentioned above, when validating the model, the time taken by the network to predict all trajectories is a performance indicator to measure the timeliness of the network's predictions. In this paper, ADE and FDE are both measured in kilometers, and the predicted average time (PAT) is measured in milliseconds.

*D. Training Parameters*

The trajectory information (the value of x, y, and z of the fighter under the Cartesian three-dimensional coordinate system) of the fighter at 8 moments in history is used as an input to predict the trajectory information at 8 moments in the future. The Adam optimization algorithm is used to optimize the weights, and the initial learning rate is set to $10^{-4}$. To avoid the problem that the learning rate is a fixed value that leads to unstable training or slow convergence, this paper adopts the learning rate decay method to carry out the learning rate, and its initial decay rate is set to 0.5, the Batch-size is set to 64, and the num_works is set to 8, which is used to realize the multi-process work, and use the GPU environment to complete the training and testing.

*E. Loss Function*

In this paper, we use the L2 paradigm as the loss function of the model with the following formula:

$$L = \sum_{i=1}^{n}(y_i - f(x_i))^2 \quad (13)$$

where $y_i$ denotes the set of coordinates of the real flight trajectory of fighter $i$ and $f(x_i)$ is the set of coordinates of the predicted trajectory points output by the model.

*F. Experimental Results*

The loss convergence curves of the CNN-LSTM network with attention mechanism and social-pooling on 100 epochs are obtained under the above network description. After the decreasing learning rate, this network converges the loss to near 0 within 20-40 epochs, which proves that the network is adequately trained.

*G. Ablation Experiments*

In this paper, 2 sets of ablation experiments are conducted to verify the effectiveness of the attention mechanism and social-pooling module on this network. The first is to remove the attention mechanism module, and the second is to remove the social-pooling module, and the networks from the two sets of experiments are compared with the complete network in the experiments respectively. The role of these two modules for the CNN-LSTM network with attention mechanism and social-pooling is verified from both qualitative and quantitative analysis perspectives. In the following, CNN-LSTM+A+SP denotes CNN-LSTM with attention mechanism and social-pooling, CNN-LSTM+A denotes CNN-LSTM with attention mechanism and without social-pooling, CNN-

LSTM+SP denotes CNN-LSTM with social-pooling and without attention mechanism.

For qualitative analysis, this paper visualizes the prediction results of different networks, including the shape, length, and direction of the trajectories.

For quantitative analysis, this paper evaluates the performance of different networks by calculating their prediction errors.

Firstly, this paper conducts a comparative experiment on whether the networks include the attention mechanism or not. From the perspective of qualitative analysis, this paper shows the predicted trajectories of the outputs of the two networks under the same inputs, to visually compare the differences in the prediction performance of the two networks, as shown in Fig. 4. The red line is the historical flight trajectory of the fighter, the blue line is the real flight trajectory of the fighter in the future moment, and the green dashed line indicates the future flight trajectory obtained from the prediction based on the historical trajectory and other information, and all the units of the coordinates in the figure are meters. (a) and (c) show the predicted trajectories of the CNN-LSTM+A+SP without and with mutation points, respectively. (b) and (d) show the predicted trajectories of the CNN-LSTM+SP without and with mutation points, respectively.

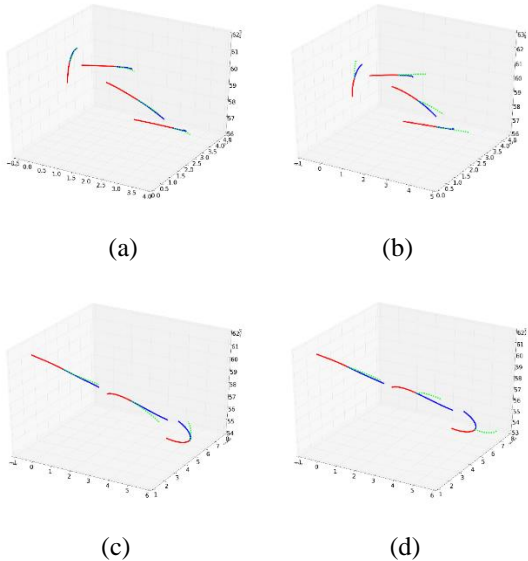

(a) (b)

(c) (d)

Fig. 4 Trajectory Prediction Results of Two Networks in Two Scenes

As shown in Fig. 4(a) (b), both networks can predict the trajectory information within the next 8 moments well in the simple straight flight condition. This is reflected in the figure by the high degree of affixing the green dashed line to the blue solid line. However, when there is a deviation in the figure, the CNN-LSTM+SP still has a large error although the trajectory trend is relatively consistent with the real trajectory. On the other hand, the CNN-LSTM+A+SP can better fit the future real trajectory whether it is a straight line trajectory or a curved trajectory with lower curvature. Therefore, the CNN-LSTM+A+SP shows better performance compared to the CNN-LSTM+SP.

As shown in Fig. 4(c) (d), there are curves with high curvature in both figures, which are reflected in the air combat scene as intense fighting and the presence of mutated trajectories. In this scene, the CNN-LSTM+SP extracts fewer spatial features from the historical trajectories, and is not able to effectively capture the mutation information present in the trajectories, and thus is not able to effectively predict the future position information. However, the CNN-LSTM+A+SP can better aggregate the features of the spatial dimension in the historical trajectories, and can better extract the inflection points in the trajectories, thus better predicting the curved trajectories with large curvature. It can be seen that the attention mechanism makes the network pay more attention to the inflection point part of the input sequence, which substantially improves the performance of the network.

From the perspective of quantitative analysis, this paper records the ADE and FDE in the above two scenes, as shown in Table 1:

Table 1 Comparison table of ablation experiments with and without attention mechanism in two scenes

| Models | Scene | ADE | FDE |
| --- | --- | --- | --- |
| CNN-LSTM+A+SP | Scene1 | 0.134 | 0.144 |
| CNN-LSTM+A+SP | Scene2 | 0.396 | 0.525 |
| CNN-LSTM+SP | Scene1 | 0.382 | 0.628 |
| CNN-LSTM+SP | Scene2 | 0.713 | 0.993 |

In summary, when the attention mechanism is introduced in the network, it is more sensitive to the mutation points in the trajectories and can better extract the features in the spatial dimension and thus better predict the complex trajectories as compared to the network without this module. On the temporal level, although the attention mechanism is introduced, it is parallel to the social-pooling in terms of network architecture, and thus the two network prediction times are similar. In this paper, the trained network was tested for trajectory prediction time-consuming with the same input variables. Record the average time it takes for each group of networks to predict the future 8 moments, and also record the ADE and FDE, as shown in Table 2:

Table 2 Comparison table of ablation experiments with and without attention mechanism

| Models | ADE | FDE | PAT |
|---|---|---|---|
| CNN-LSTM+A+SP | 0.235 | 0.388 | 6.065 |
| CNN-LSTM+SP | 0.497 | 1.035 | 6.055 |

Meanwhile, this paper conducts comparative experiments on whether the networks include the social-pooling or not. Firstly, from the perspective of qualitative analysis, this paper shows the predicted trajectories of the two sets of network outputs under the same input, as shown in Fig. 5, where the meanings represented by the curves remain the same as in Fig. 4.

Fig. 5 (a) shows the prediction trajectory of the CNN-LSTM+A+SP. Fig. 5 (b) shows the prediction trajectory of the CNN-LSTM+A.

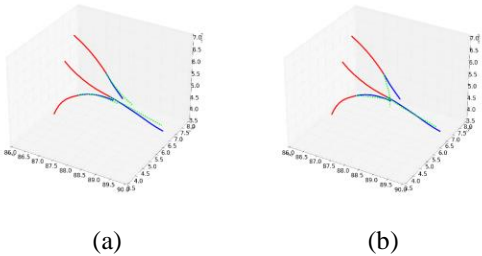

(a)　　　　　　　(b)

Fig. 5 Trajectory Prediction Results of Two Networks for the Same Scene

As shown in the lower two trajectories in Fig. 5(a)(b), both networks can better predict the future trends of simple straight trajectories and curved trajectories with higher curvature through the attention mechanism. However, the network without the social-pooling cannot better aggregate the spatial information features between different fighters, resulting in a relatively large error between the predicted trajectories and the real trajectories for the uppermost curve in Fig. 5 (b).

From a quantitative analysis point of view, this paper records the ADE and FDE for the above two scenes as shown in Table 3:

Table 3 Comparison table of ablation experiments with and without the social-pooling in the same scene

| Models | Scene | ADE | FDE |
|---|---|---|---|
| CNN-LSTM+A+SP | Scene1 | 0.274 | 0.397 |
| CNN-LSTM+A | Scene1 | 0.326 | 0.451 |

In summary, when the social-pooling is introduced in the network, it can better aggregate the interaction information between the fighters in the spatial dimension, and thus better predict the trajectories, compared to the network without this module. In the temporal dimension, although social-pooling is introduced, it is parallel to the attention mechanism in the network architecture, so the two networks have similar prediction times. In this paper, the same test method as above is used to record the ADE and FDE of the two networks, and the time consumed by the networks is also recorded, as shown in Table 4:

Table 4 Comparison table of ablation experiments with and without the social-pooling

| Models | ADE | FDE | PAT |
|---|---|---|---|
| CNN-LSTM+A+SP | 0.235 | 0.388 | 6.065 |
| CNN-LSTM+A | 0.347 | 0.491 | 6.027 |

Based on the above 2 sets of ablation experiments, this chapter verifies that the attention mechanism and the social-pooling play important roles in the improvement of the trajectory prediction accuracy of the CNN-LSTM network from both qualitative and quantitative analysis perspectives.

## IV. CONCLUSION

In this paper, we introduce the difficulty of the trajectory prediction problem, apply CNN-LSTM network with attention mechanism and social-pooling to the fighter flight trajectory prediction task, and jointly verify the important role of the attention mechanism and the social-pooling for this network through the perspective of qualitative analysis and the perspective of quantitative analysis. Through experiments, it is verified that the attention mechanism and social-pooling enable the error between the predicted trajectory and the real trajectory to be smaller.